\documentclass[sigconf]{acmart}

\renewcommand\footnotetextcopyrightpermission[1]{} 
\usepackage{graphicx}
\usepackage{subfigure}
\usepackage{epstopdf}
\usepackage{multirow}
\usepackage{multicol}
\usepackage{arydshln}
\usepackage[linesnumbered,boxed,ruled,commentsnumbered]{algorithm2e}
\usepackage{mathrsfs}
\usepackage{amsthm}

\usepackage{amsmath}
\let\oldnl\nl
\newcommand{\nonl}{\renewcommand{\nl}{\let\nl\oldnl}}

\setcopyright{none}

\begin{document}

%
\title{ALCNN: Attention-based Model for Fine-grained Demand Inference of Dock-less Shared Bike in New Cities}
\author{Chang Liu}
\email{only-changer@sjtu.edu.cn}
\affiliation{%
  \institution{Shanghai Jiao Tong University}
  \streetaddress{800 Dongchuan Road}
  \city{Shanghai}
  \country{China}
}
\author{Yanan Xu}
\email{xuyanan2015@sjtu.edu.cn}
\affiliation{%
  \institution{Shanghai Jiao Tong University}
  \streetaddress{800 Dongchuan Road}
  \city{Shanghai}
  \country{China}
}
\author{Yanmin Zhu}
\email{yzhu@sjtu.edu.cn}
\affiliation{%
  \institution{Shanghai Jiao Tong University}
  \streetaddress{800 Dongchuan Road}
  \city{Shanghai}
  \country{China}
}
\begin{abstract}
In recent years, dock-less shared bikes have been widely spread across many cities in China and facilitate people's lives. However, at the same time, it also raises many problems about dock-less shared bike management due to the mismatching between demands and real distribution of bikes. Before deploying dock-less shared bikes in a city, companies need to make a plan for dispatching bikes from places having excessive bikes to locations with high demands for providing better services. In this paper, we study the problem of inferring fine-grained bike demands anywhere in a new city before the deployment of bikes. This problem is challenging because new city lacks training data and bike demands vary by both places and time. To solve the problem, we provide various methods to extract discriminative features from multi-source geographic data, such as POI, road networks and nighttime light, for each place. We utilize correlation Principle Component Analysis (coPCA) to deal with extracted features of both old city and new city to realize distribution adaption. Then, we adopt a discrete wavelet transform (DWT) based model to mine daily patterns for each place from fine-grained bike demand. We propose an attention based local CNN model, \textbf{ALCNN}, to infer the daily patterns with latent features from coPCA with multiple CNNs for modeling the influence of neighbor places. In addition, ALCNN merges latent features from multiple CNNs and can select a suitable size of influenced regions. The extensive experiments on real-life datasets show that the proposed approach outperforms competitive methods.
\end{abstract}

%
%



%

%

%
\maketitle

\section{Introduction}\label{sec:intro}

    Recently, dock-less shared bike services have achieved great success and reinvented bike sharing business in China, especially in major cities.
    Dock-less shared bikes provide an environmentally friendly solution to solve the last mile problem which refers to the troublesome distance between home and the nearest traffic center. Many dock-less shared bike companies have grown rapidly and seized the market, such as ofo and Mobike\footnote{\url{https://mobike.com/cn/}}.
    The rules they set are similar: user can find a dock-less shared bike via a GPS-based smart phone APP and follow the APP's instructions to unlock the bike (scan the QR code or enter passwords), then ride the bike to anywhere they want, finally lock the bike and pay some money to the company. The convenience of this mode makes many people benefit from it and reinvents bike industry in China.
\begin{figure}[tp]
    \centering
    \includegraphics[width = 0.45\textwidth]{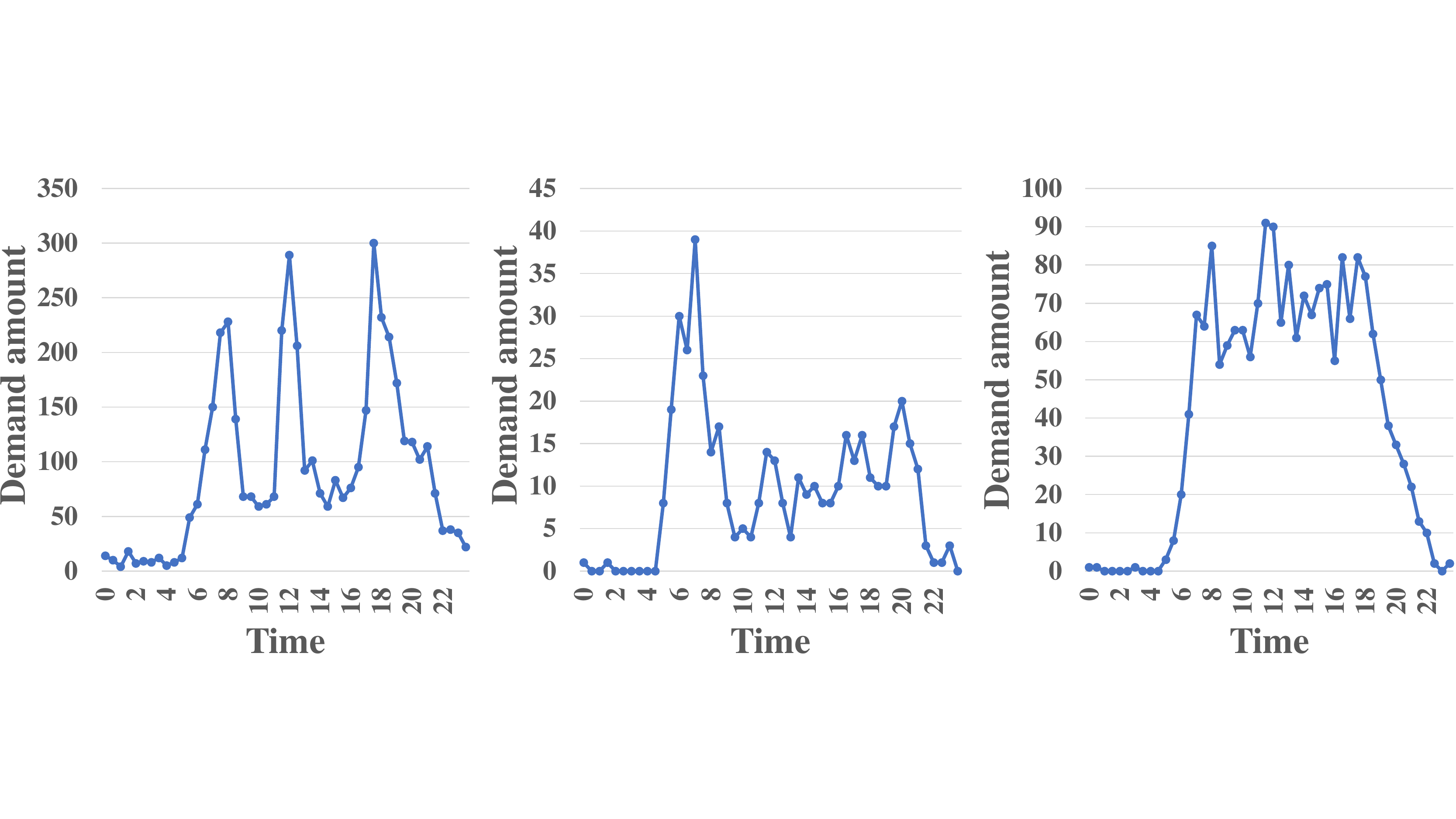}
    \caption{Fine-grained bike demands in three different places in a same day. Bike demands are counted for each half hour.}
    \label{fig:bikeDemandDiff}
\end{figure}

    But the prosperity of this new way of transportation will inevitably lead to new problems. Before companies officially deploy dock-less shared bikes in a new city, they need to make a plan of managing bikes as the real-time distribution of bikes may not match bike demand. Manage bikes according to bike demands will greatly improve the effective use of bikes, i.e., serving more costumers with less cost. However, the bike demands in a city vary spatially and temporally. We regard the curve of demand per half hour in one area as \textbf{fine-grained bike demand}. Figure~\ref{fig:bikeDemandDiff} shows fine-grained bike demands in a day for three locations.
    We can see that for each location, its fine-grained bike demand varies with time. Different places have different bike demands. If we know the fine-grained bike demands of the three places in Figure~\ref{fig:bikeDemandDiff}, at noon we can move spare bikes in the second place to the first place and the third place.

    
    The fine-grained bike demands are easy to collect in a city which has deployed shared bikes for a while. But for a city having no shared bikes, the lacking of historical bike demand records makes it hard to build a prediction model. 
    In this paper, we focus on a problem of inferring fine-grained bike demands within a time slot for a place in new cities which have not deployed dock-less shared bikes. We want to build an demand inference approach based on geo-related data, such as POI, road and transportation, in a city having bikes. Then we can transfer it to a new city and infer bike demands. Companies can use the inference result to design a schedule algorithm before the deployment to balance the supply and demand of bikes in all regions~\cite{li2018dynamic}.

    Many existing works on dock-less shared bike system take geo-related data into consideration to infer the distribution of bike demand in a city~\cite{liu2018inferring,liu2018will}. But they only focused on the spatial distribution and omitted temporal variability which is also important for the management of bikes. In this paper, we aim to infer the bike demand distribution over both spatial and temporal dimensions.

    However, there are three challenges for solving the problem. First, the bike demands are time-varying but we only have static geographic data like POIs. The bike demands vary with time in a day for one place. Even for the same time of different days, their demands may also vary a little. 
    Second, our task is to infer fine-grained bike demands in a new city having not deployed shared bikes and we don't have any bike demand data in that city. The data distribution may be different in diverse cities.
    Finally, each place is influenced by their neighbors and the size of area influenced by neighbors is hard to determine.

    To address these challenges, we first analysis the fine-grained bike demand data and find that there exist daily demand patterns for many places (details are shown in Section~\ref{sec:dataAnalysis}). We propose to utilize discrete wavelet transform (DWT) to mine daily bike demand patterns from fine-grained bike demands of every day in each place. The mined daily demand patterns are stable and are used as the ground truth of inference.
    To deal with the problem of lacking of training data in a new city, we next utilize multi-source geographic data to extract discriminated features and train a inference model to predict bike demands. The inference model is trained with data from an old city which has deployed shared bikes and transferred to new cities. We employ correlation Principal Component Analysis (coPCA) to realize the distribution adaption when extracting latent features from geographic data of both old city and new city.
    For considering the influence of neighbors, we then propose an attention-based model, ALCNN, which utilizes CNN to aggregate geographic features from neighbors in a local region for each place. As the size of influenced area of different places varies, we build several CNN models with local regions of different sizes and merge their learned latent features using attention mechanism. In other words, the attention mechanism can choose local regions of a suitable size for each place automatically.

    The main contributions of this paper are summarized as follows.
    \begin{itemize}
        \item To the best of our knowledge, we are the first to study the problem of inferring fine-grained bike demands in a new city. We present the importance of finding temporal patterns from fine-grained bike demands and propose to use discrete wavelet transform to mine daily patterns from bike demands.
        
        
        \item We utilize local CNN to aggregate geographic features of neighbors in local regions of various sizes for each place. We propose to use attention mechanism to merge the learned latent features from local regions of different sizes. 
        
        \item We evaluate our inference approach on real Mobike data of three cities in China. The results show that our approach outperforms all baselines. The attention mechanism can improve the inference performance.
    \end{itemize}
    
    The remainder of this paper is organized as follows. In Section~\ref{sec:overview}, we formulate our fine-grained bike demand inference problem and introduce the framework of our approach. In Section~\ref{sec:dataAnalysis}, we analysis the dataset of dock-less shared bikes of Mobike. Next, we introduce feature extraction processes and our proposed model ALCNN in Section~\ref{sec:featureExtract} and Section~\ref{sec:model}, respectively. Then, we conduct extensive experiments to evaluate our approach in Section~\ref{sec:experiment}. 

\section{OVERVIEW}\label{sec:overview}

    In this section, we first introduce several fundamental definitions and present the formulation of our bike demand inference problem.
    Next, we show the overall framework of our approach.

\subsection{Preliminary}

    \begin{definition}[\textbf{City grid $g$ and city grid map $G$}]
    We regard a city as one rectangle and divide it into $n \times m$ disjointed grids of the same size according to latitude and longitude. Each grid is denoted by $g_{i,j}$, where $1 \leq i \leq n$ and $1 \leq j \leq m$. As a result, the city can be seen as a grid map, i.e., $G = \left\{g_{i,j} \right\}$. For convenience, we use function $inside(p,g)$ to judge whether a point $p$ is inside a grid $g$.
    \end{definition}
    
    \begin{definition}[\textbf{Mobike trip record collection $R$}]
        We use the data from a dock-less shared bike company called Mobike. Our dataset of dock-less shared bikes is a collection of riding trip records, i.e., $R=\{r\}$. Each record is a tuple $r=(l^s, t^s, l^e, t^e)$ the elements of which denote the starting location and time, ending location and time, respectively. One location $l$ is formed by its longitude and latitude. 
    \end{definition}
    If the starting location $l^s$ of one record $r$, is within a grid $g$, we say that a renting behavior happened in that grid. Based on the trip records, we define a bike demand set $D$ as follows.
    
    \begin{definition}[\textbf{Fine-grained bike demand set $D$}]\label{def:demandSet}
        For easy processing, we divide each day to into $k$ time slots of the same length and use $(x, h)$ to denote the $h$-th time slot in the $x$-th day. Next, we can obtain fine-grained bike demands of each grid in a certain time slot with the following equation.
        \begin{equation}
            N^{x,h}_{g} = \left| \{ r| r \in R, inside(r.l^s,g), r.t^s \in (x,h) \}  \right|.
        \end{equation}     
        
    As we aim to infer daily bike demand pattern of each grid, we use one demand vector $d^x_g$ to denote bike demands of all $k$ time slots in day $x$ for grid $g$, i.e., $d_{g}^{x} =  [N^{x,1}_{g},N^{x,2}_{g},\cdots,N^{x,k}_{g}]$. For each grid, its bike demand vectors of different days can form a set of fine-grained bike demands $D_{g} = \{ d_{g}^{x} \}$
    \end{definition}

\begin{figure}[t]
    \centering
    \includegraphics[width = 0.5\textwidth]{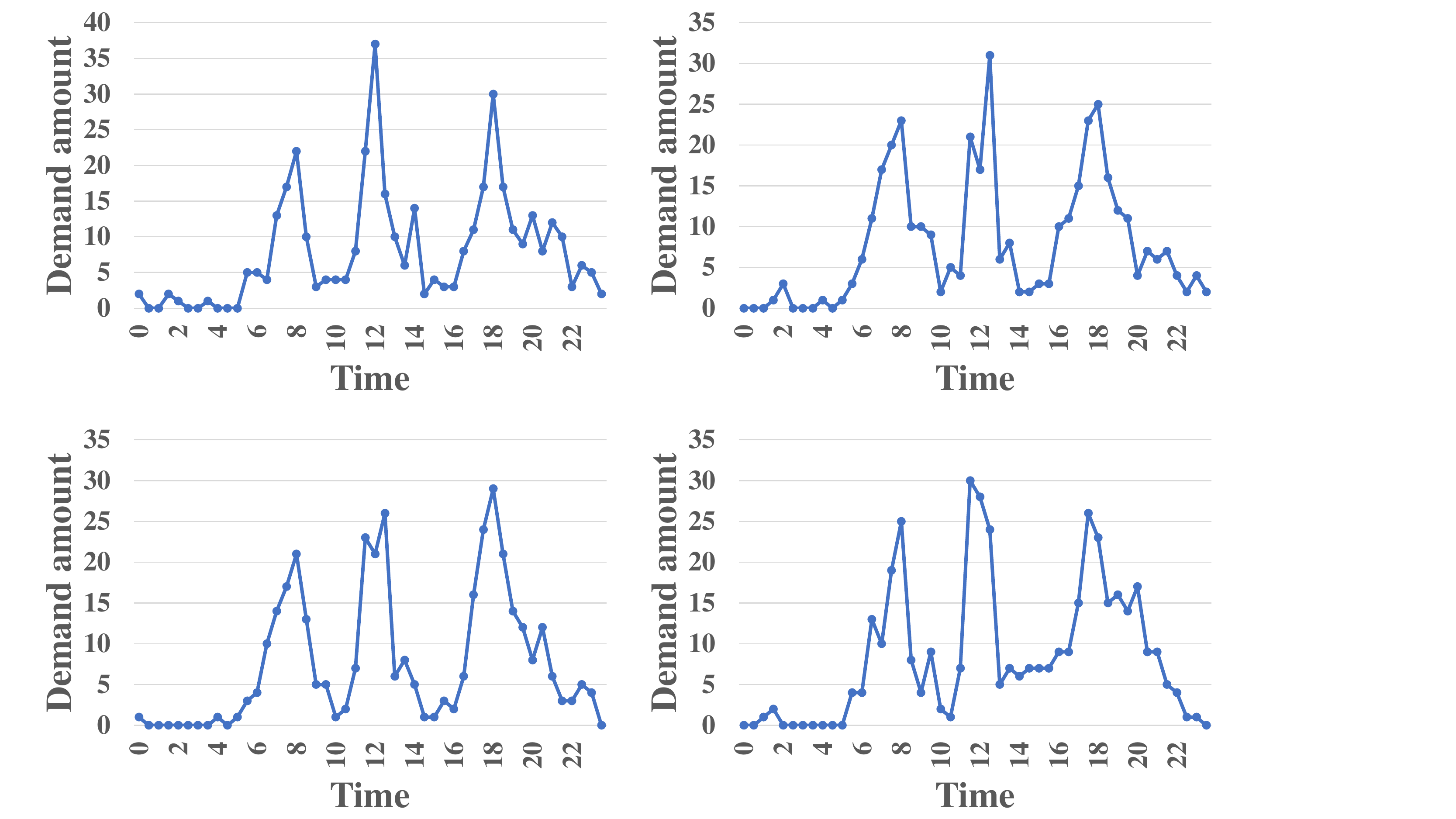}  
    \caption{Fine-grained bike demands on different days in a same grid}
    \label{fig:demandDiffSameGrid}
\end{figure}

   As Figure~\ref{fig:demandDiffSameGrid} shows, we find that in most grids, their fine-grained bike demands in different days are similar to each other. Based on the observation, we can draw two conclusions. On the one hand, most grid has their daily patterns among the bike demands. On the other hand, it should be better to mine the daily patterns which will be more stable and benefit the convergence of our inference model. Therefore, we define daily temporal patterns of bike demand for each grid as follows.
   
\begin{figure*}[pt]
    \centering
    \includegraphics[width = 0.7\textwidth]{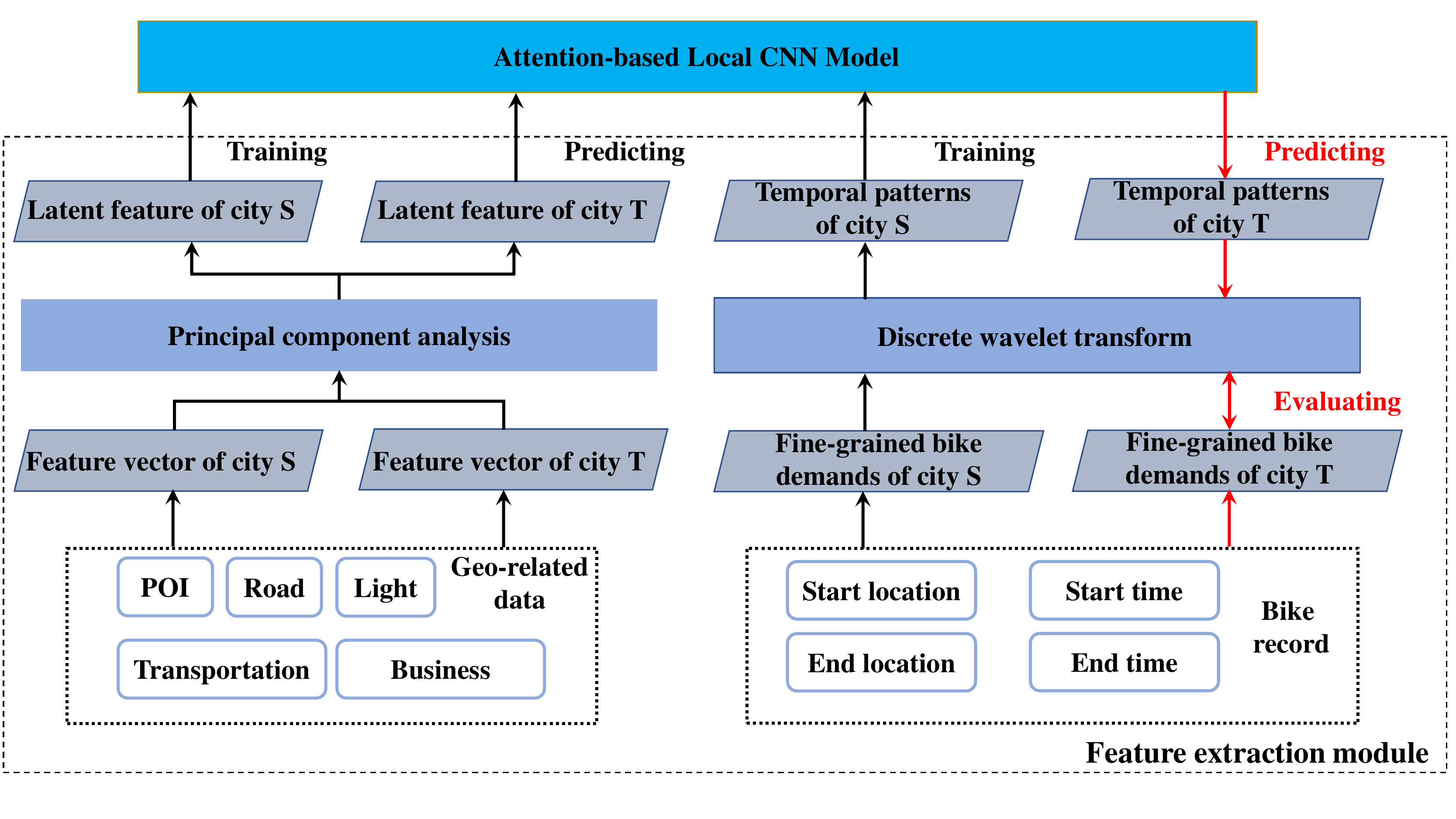}  
    \caption{The framework of our system}
    \label{fig:modelStruc}
\end{figure*}
    
    \begin{definition}[\textbf{Daily demand pattern $P$}]
        For each grid, we aim to find a demand vector $d_{g}$ which is similar to all demand vectors of most days. Considering that the bike demands of different grids or different days can vary widely, we normalize all demand vector with its summation of all elements. 
        We employ Kullback-Leibler(KL) divergence to measure the difference between demand vectors. If we manage to find a demand vector $d_{g}$ and the KL divergence $d_{g}$ and all other demand vectors of the same grid is smaller than a threshold $\beta$, we treat $d_{g}$ as the daily demand pattern $P_{g}$ of grid $g$. Note that $d_{g}$ does not have a superscript $x$ because $d_{g}$ is not a real bike demand vector of any day. We will talk about how to find $d_{g}$ by DWT latter.
        \begin{equation}\label{eq:demandPattern}
            P_{g} = d_{g}, \ KL(d_{g} || d^x_{g}) < \beta, \forall d^x_{g} \in D_g
        \end{equation}
    \end{definition}



    
    \textbf{Problem Formulation.}
    We consider two cities S and T, where S has deployed dock-less shared bikes and T is a new city to launch shared bike business. We divide both cities into grids of the same size. We regard city S as a source domain while city T as a target domain. Given a set of raw bike data in city S, we compute the set of fine-grained bike demands D according to Definition~\ref{def:demandSet}. Our goal is to infer fine-grained bike demands in the new city T, based on the known fine-grained bike demands $D$ from city S and multiple geo-related data (e.g., POI and road networks) in both city S and T. 

\subsection{Inference Framework}

    To achieve the fine-grained bike demand inference goal, we propose an inference system consisting of two major components, i.e., feature extraction and model training, as shown Figure~\ref{fig:modelStruc}. 

    In the feature extraction component, we first extract features from multi-source geographic data including POIs, road networks, satellite lights, transportation centers and business centers. We also process the original records of Mobike and get bike demands. Next, we utilize Correlation Principal Component Analysis (coPCA) and Discrete Wavelet Transform (DWT) to deal with the two kinds of features in a further step. coPCA produces discriminative latent features as the input of our inference model. coPCA processes geo-related data in source city and target city together. It realizes the distribution adaption between the two cities and improves the inference performance. DWT extracts temporal patterns from bike demand vectors and can improve the stability of our inference model as it represents the demand vectors with fewer dimensions and parameters.

    In the second component, we feed the extracted latent features to a proposed attention-based local CNN model. We train the model with data of source city S and transfer the trained model to a new city T. Using the geographic features of the target city T, we can infer the bike demand in the new city. The proposed attention-base local CNN model can select features of each grid and its neighbors and processes them with multiple convolution networks for modeling influence of neighbors. The attention mechanism in the model can help choosing the most suitable size of local region to provide better inference results.  




\section{DATA ANALYSIS}\label{sec:dataAnalysis}








    Before we conduct empirical analysis on dock-less shared bike data, we need to set a evaluation method to define how close does two different fine-grained demands are. In order to do this, we adopt the Kullback-Leibler (KL) divergence to compute the closeness between two fine-grained bike demands. Based on the definition in Section~\ref{sec:overview}, we can define the KL divergence:

    \begin{equation}
        KL(d_{g}^{x} || d_{g}^{y}) = \sum_{t=0}^{k} N_{g}^{x,t} log \frac{N_{g}^{x,t}}{N_{g}^{y,t}}.
    \end{equation}

    Intuitively, the smaller KL divergence is, the closer two fine-grained demand vectors are. In our context, the closeness of every two consecutive bike demand vectors indicate the stability of daily fine-grained bike demands.

\begin{figure}[t]
    \centering
    \includegraphics[width = 0.45\textwidth]{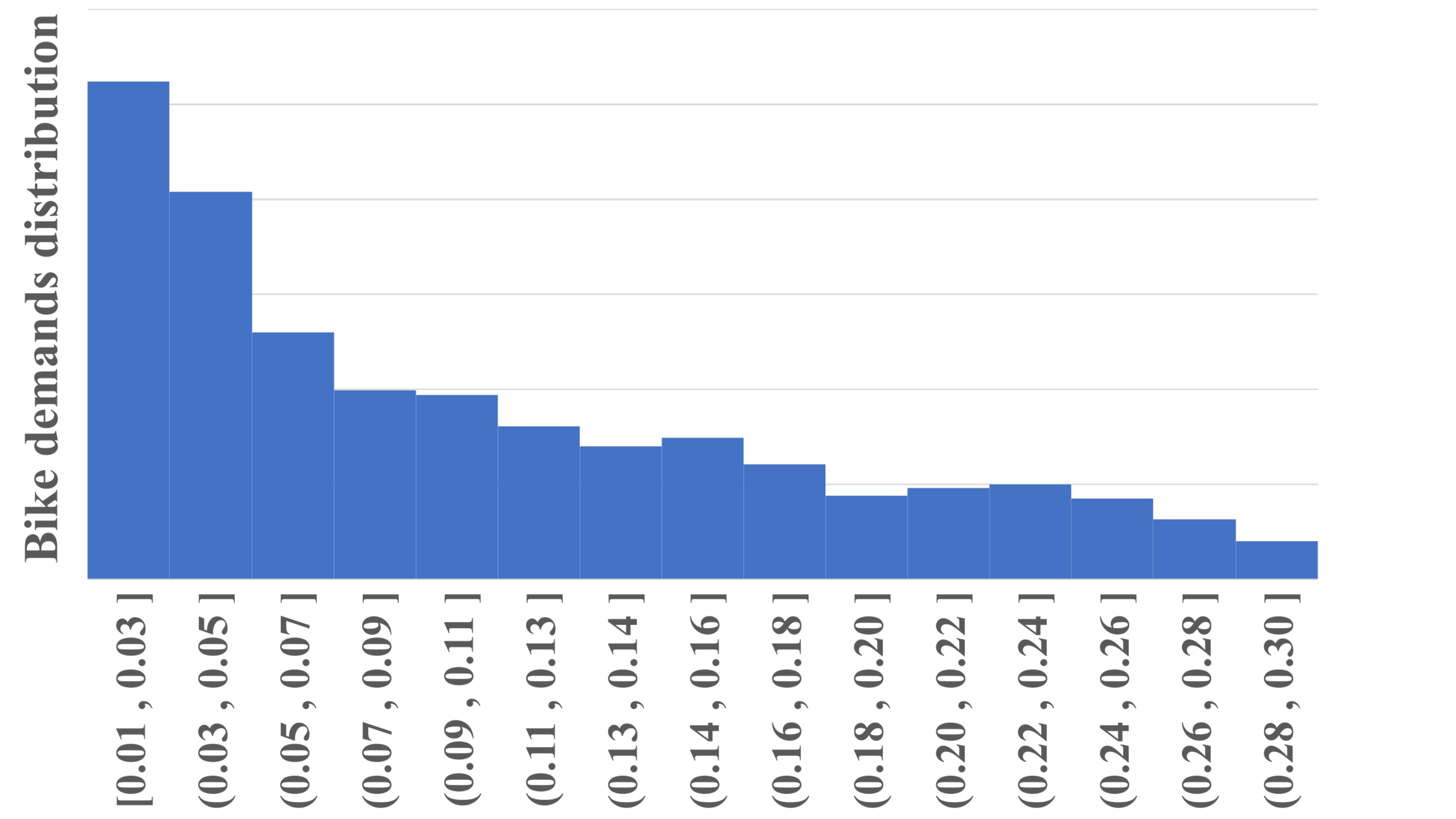}  
    \caption{Distribution of divergence between demands of different days}
    \label{data}
\end{figure}

    As we have many fine-grained bike demands on each grid for different days, we still need a way to evaluate the divergence between all fine-grained bike demands on each grid. As we mentioned in Section~\ref{sec:overview}, we use the maximum of Kl divergence among all fine-grained bike demands as evaluation. The lower divergence means the higher possibility we can find a pattern on that grid.

    \begin{equation}
        Divergence = max\ KL(d_{g}^{x} || d_{g}^{y}), \ \ x,y \in X,
    \end{equation}
    where $X$ is a vector of days. By the definition of divergence, we can do empirical analysis on real data, to see whether there will be necessary to find a pattern. We scan all the real data and get the result as shown in Figure~\ref{data}.

    We can see about $30\%$ of real fine-grained demands have a divergence less than $0.03$, and almost $70\%$ of real fine-grained demands have a divergence less than $0.11$. Based on observation, we draw a conclusion that \textbf{most grids in the city have temporal patterns}, which is necessary to mine temporal patterns for further work.
    



    Actually, temporal patterns are very complicated, for example, even if we mainly conside peaks in patterns, the amount, time and height of peaks will all be changeable in everywhere in the city. However, we don't think that the difference among temporal patterns is just random.
    Instead, we think that there is a connection between temporal patterns and the geographic data.
    For example, most grids near subway stations have ``TRIPLE PEAK'' patterns (three peaks during the day), because that subway station has a large flow of people and it comes to a burst on every possible peak, while the grids near business center always have ``DOUBLE PEAK'' patterns (two peaks during the day), for that the people working in business center have a rapid pace of working, hence they don't have enough time to use bikes in noon, which makes the peak at noon does not exist.

    Above all, we find that there is a need to find temporal patterns from fine-grained bike demands, for that most grids have low divergence. We also find that temporal patterns are typically different because of geographic diversity, which inspires us to leverage geographic data sources such as POI, subway station, business center when inferring the patterns.

\section{Feature Extraction}\label{sec:featureExtract}

    In this section, we will show the details about feature extraction component. The result of feature extraction will be the input of our propose ALCNN model in section~\ref{sec:model}.

\subsection{Features from Multi-source Data}

    Geographic data is utilized to infer bike demands in this paper as they reflect the condition of transportation and high related to the bike demands. In this paper, we consider $5$ kinds of geographic data including POI, road networks, nighttime light, transportation centers, and business centers. We will give a brief introduction to each kinds of data and present the feature extraction methods. The extracted detailed features from every kinds data are denoted by $F_{p}$, $F_{r}$, $F_{s}$, $F_{t}$, and $F_{b}$, respectively.

    \textbf{1.POI Features}: $F_{p}$
    
    Each POI represents a city venue with its name, address, category and spatial coordinates.
    The number and diversity of POIs on a grid reflects its prosperity and hence are related to bike demand density. We extract the following three POI features for each grid:

    (1) POI category frequency ($\vec{P}$): We associate each city grid $g$ with a POI category frequency
    vector $\vec{P}$. $\vec{P}$ is a $17$-dimension vector and the $i$-th element in $\vec{P}$ indicates the number of venues of the $i$-th category located in the grid $g$.
    
    (2) Number of POIs ($P_{num}$): We can count the total number of POIs $P_{num}$ based on frequency vector $\vec{P}$ on a grid: $P_{num} = \sum_{i}\vec{P_{i}}$.
    
    (3) POI entropy ($P_{en}$): Besides the number of POIs, we also compute the entropy $P_{en}$ based on $\vec{P}$ for indicating the heterogeneity of POIs in a grid.
    \begin{equation}
        P_{en} = -\sum_{i}\frac{\vec{P_{i}}}{P_{num}} \times log\frac{\vec{P_{i}}}{P_{num}}.
    \end{equation}
    
    \textbf{2. Road Network Features}: $F_{r}$
    
    Intuitively, the more roads located in one grid, the more convenient the traffic will be. 
    In our dataset, each road consists of its name, category (or level), start point and end point.
    We propose to use two road network features for every grid: 

    (1) Road category frequency ($\vec{R}$): We can associate road category frequency vector $\vec{R}$ for each city grid $g$.  
    The $i$-th element in $\vec{R}$ denotes the number of $i$-th type roads overlapped with grid $g$ .
    
    (2) Number of roads ($R_{num}$): After we get the frequency vector $\vec{R}$, we can count the total number of overlapped roads $R_{num}$: $R_{num} = \sum_{i}\vec{R_{i}}$.
    
    \textbf{3. Nighttime Light Features}: $F_{s}$
    
    The nighttime light data are collected by satellites. We get the light intensity by sample points for every $50$ meters on the map.
    Intuitively, nighttime light intensity are positively correlated with business prosperity and population density of a grid which implies large shared bike demands. We identify two kinds of features:
    
    (1) Average light intensity ($S_{a}$): To reduce the influence of data noises, we calculate the average light intensity. Denote $S$ as the set of light points in the city, and $I$ as the intensity of every point. We compute the average light intensity $S_{a}$ in $g$:
    \begin{equation}
        S_{a}(g) = \frac{\sum_{i \in S(g)}I(i)}{|S(g)|}.
    \end{equation}
    
    (2) Distance to the nearest light centre ($S_{dis}$):  From the map of nighttime lights, we identify that there are several centers whose light intensity is large than other places. Each light center can be denoted by their longitude and latitude. Based on the light centers, we compute the geographic distance between each grid and their nearest light center.
    \begin{equation}
        S_{dis}(g) = \min \{\text{distance}(g, c)| \ c \in S_c\},
    \end{equation}
    where function $\text{distance}(x,y)$ is the function to calculate geographic distance between two location $x$ and $y$. For a grid $g$, we use the longitude and latitude of its center as the location. $S_c$ denotes the set of all light centers in a city.

    \textbf{4. Transportation Features}: $F_{t}$
    
    As we all know, most dock-less shared bikes are parked near the transportation centers, such as subway station. Because people usually get off buses or metros in transportation centers and need to find a bike for traveling in a short distance, e.g., from bus station to home.
    We can extract two main features:
    
    (1) Number of transportation centers ($T_{num}$):
    We use $T$ as the set of transportation centers in the city. Then, we can count the number of transportation centers in each grid: 
    \begin{equation}\label{eq:tnum}
      T_{num}(g) = | \left\{i\ |\ inside(i,g) , i\in T\right\}|
    \end{equation}
    
    (2) Distance to the nearest transportation centers ($T_{dis}$):
    For each grid, we also compute its center to the most nearest transportation centers:
    \begin{equation}
        T_{dist}(g) = \min \{ \text{distance}(g, i) | i \in T \}
    \end{equation}

    \textbf{5. Business centre features}: $F_{b}$
    
    Around the business centers, the flow of people can be very large which indicates large bike demands.
    In our business centre dataset, each business centre is denoted by its name, level, and location. We use $B$ to denote the set of all business centers in a city. We extract two business centre features for each grid:
    
    (1) Distance to the nearest business centre ($B_{dis}$)
    
    We use $B$ as the set of all business centres, we have:
    \begin{equation}
        B_{dis}(g) = \min \{\text{distance}(g, i)| \ i \in B\}
    \end{equation}
    
    (2) Level of the nearest business centre ($B_{level}$)

\begin{equation}
	B_{level}(g) = (\mathop{\arg\min}_{i} \ \text{distance}(g, i).level, \ \  i \in B
\end{equation}

\subsection{Correlation Principal Component Analysis for Transfer Learning}

    As our task is to infer fine-grained bike demands in a new city and the distribution of two cities may be different, we propose to apply coPCA over the extracted features to achieve distribution adaption. 
    Originally, PCA aims to use latent features with low dimension to represent the raw features. If each data sample is seen as one point in a coordinate system, PCA rotates the coordinate axes and on the top several axes the samples have the largest variance. The latent features on these axes can reserve effective information and filter out data noises. 
    In this paper, we use coPCA to find a transformation to minimize difference between the distributions of data from two cities.

    We first construct two matrices to store features of grids from two cities, respectively. In the matrices, each row denotes one grid in a city and every column stands for one kind of feature, in other words the matrices are concatenations of $F_p,F_r,F_s,F_t,F_b$ and $F_p^{'},F_r^{'},F_s^{'},F_t^{'},F_b^{'}$. Next, we concatenate the two matrices in the first dimension and apply PCA to the entire matrix. When we get the result of coPCA, we will divide the output matrix $H$ into two parts $F, F^{'}$ along the first dimension. The two parts can be seen as the results of dimension reduction of data from source city and target city, respectively. 



    
    

\begin{figure*}[t]
    \centering
    \includegraphics[width = 0.7\textwidth]{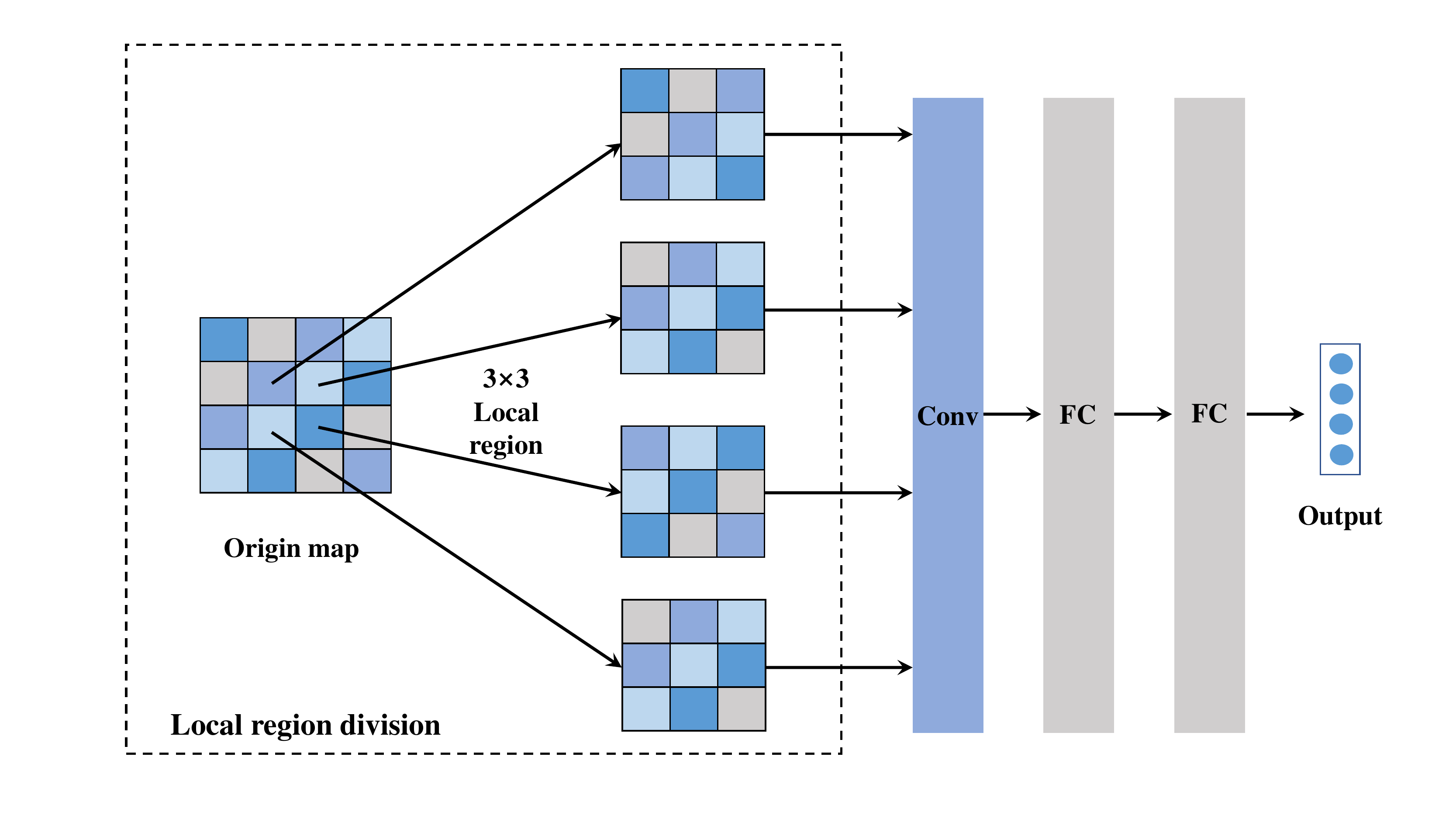}  
    \caption{The structure of our local CNN model. Four grids are selected as training samples. The size of local region of each grid is set to $3 \times 3$ in this figure.}
    \label{fig:localCNNStruc}
\end{figure*}

\subsection{Discrete Wavelet Transform}

    As we discuss in Section~\ref{sec:dataAnalysis}, each grid has its daily bike demand patterns. Moreover, the dimension of fine-grained bike demand vector can be large which may lead to high computation complexity and a large number of parameters. Inspired by the dimension reduction of input features, we also want to reduce the dimension of output, i.e., bike demands.
    We utilize Discrete Wavelet Transform(DWT) to solve the problem. As with other wavelet transforms, a key advantage it has over Fourier transforms is temporal resolution: it captures both frequency and location information (location in time).

    The process of DWT is actually passing through a series of filters. For convenience, we use two filters, the low pass filter and high pass filter, which are decided by mother wavelets and are known as quadrature mirror filter\footnote{\url{https://en.wikipedia.org/wiki/Quadrature_mirror_filter}}. According to early researches on DWT\cite{kronland1987analysis}, the low pass filter can filter the original signal to the approximation coefficients, while the high pass filter can filter the original signal to the detail coefficients. The main reason why we use DWT is that DWT can obtain the approximation by low pass filter, which uses fewer parameters because of the other parameters are in high-frequency information. The basic math formula is as follow:
    
    \begin{align}
        y_{low}^{x}[n]  & = \sum_{k}d_{g}^{x}[k]l[2n - k],\\
        y_{high}^{x}[n] & = \sum_{k}d_{g}^{x}[k]h[2n - k],\\
        d_{g} &= \frac{\sum_{x \in X} idwt(y_{low}^{x})}{|X|},
    \end{align}
    where $d_{g}^{x}$ is the fine-grained bike demand in grid $g$ on day $x$. $n$ is the reduced parameter size, which is half of origin size in this case. $h$ is high-pass filter while $l$ is low-pass filter. $y$ is the result after DWT. Then we use inverse DWT (idwt) to transfer $y_{low}$ to a new fine-grained demand, and we ensemble all new fine-grained demands in grid $g$ on different days $X$ to get the candidate daily pattern $d_{g}$. The selection of $l$ and $h$ is based on the mother wavelet, which is not unique, like Haar wavelets, Daubechies wavelets, Symlets wavelets. In this paper, we choose Daubechies wavelets to do our work.

    Since we have found a way to use fewer parameters to describe the origin distribution approximately, we can use DWT to preprocess a fine-grained bike demand and get a profile on every day. Then, we can get the candidate temporal pattern $d_{g}$ of one grid from all profiles assembled from its original fine-grained bike demands on different days. We need to set a standard divergence threshold $\sigma$ to judge whether $d_{g}$ is a temporal pattern or the fine-grained demands of this grid are just disorganized, which is defined in Section~\ref{sec:overview} and \ref{sec:dataAnalysis}. After all the above steps, we can get the temporal pattern on every grid or decide that it doesn't have a temporal pattern.
    
    Above all, we choose Daubechies wavelets to do DWT. For each existing fine-grained bike demand one grid $g$, we use DWT to find its profile, which will reduce the number of parameters. Then, we set a threshold $\beta$ to judge whether this grid has a temporal pattern using the similarity among all its profiles in different days. After these steps, we find most grids having temporal patterns and put them into our training set, with the mined patterns as their labels.

\section{ATTENTION BASED LOCAL CNN MODEL}\label{sec:model}

    After extracting features from multi-source data, we propose an \underline{A}ttention-based \underline{L}ocal \underline{CNN} (ALCNN) model to infer bike demand of each grid. 

\subsection{Local CNN}

    As we all know, geographically adjacent regions share similar characteristics. E.g., if a place is near a metro station, it will have high pedestrian volumes as the station. And the station has a large influence on the traffic of that place. Inspired by this thought, we develop a local convolution network to model the influence of neighbors for each grid.

    Figure~\ref{fig:localCNNStruc} shows the structure of our proposed local CNN which can be seen as a variant of traditional CNN. 
    The input of local CNN is the feature tensor of all grids in a city. We first select one target grid and its neighbors within a certain distance as a local region with Equation~\ref{eq:selectRegion}. E.g., in Figure~\ref{fig:localCNNStruc}, the target grid and its adjacent grids (distance equal to $1$) are selected. Note that every local region has all the features of the central grid itself and its neighbors, but it only has the label (temporal pattern after DWT) of the central grid, since it is just an enlargement of the central grid actually.

    \begin{equation}
    \begin{split}\label{eq:selectRegion}
        Input(g_{ij}, w) &=  F_{i-w:i+w, j-w:j+w, :}, \\
        0\leq i \leq n,  &  0\leq j \leq m,
    \end{split}
    \end{equation}
    where $g_{ij}$ is a target grid we consider, $F$ is the feature tensor for all grids in the city after PCA. The selected feature map for every grid is a $w \times w \times d^{'}$ tensor called $T_{g}$, where $w$ is the neighbor size and $d^{'}$ is the dimension of new features after coPCA.
    
    Next, convolution operations are performed on the feature map of selected local region. Different target grids share CNN parameters. Then the CNN is followed with two fully connected layers for learning interactions between features.

    \begin{equation}
        \begin{split}
            Z_{i0} &= \sigma(Conv(K_{i},T_{g}) + b_{i0}),\\
            Z_{i1} &= \sigma(W_{i1}Z_{i0} + b_{i1}),\\
            Z_{i2} &= \sigma(W_{i2}Z_{i1} + b_{i2}),\\
        \end{split}
        \label{layer}
    \end{equation}
    where $K_{i}$ is the parameter tensor of $i$-th convolution filter, $b_{i0}, b_{i1}\\, b_{i2}$ are the $i$-th bias vectors, $W_{i1}, W_{i2}$ are $i$-th weight matrices and $\sigma$ is the activation function.

\subsection{Attention Mechanism}

\begin{figure*}[t]
    \centering
    \includegraphics[width = 0.8\textwidth]{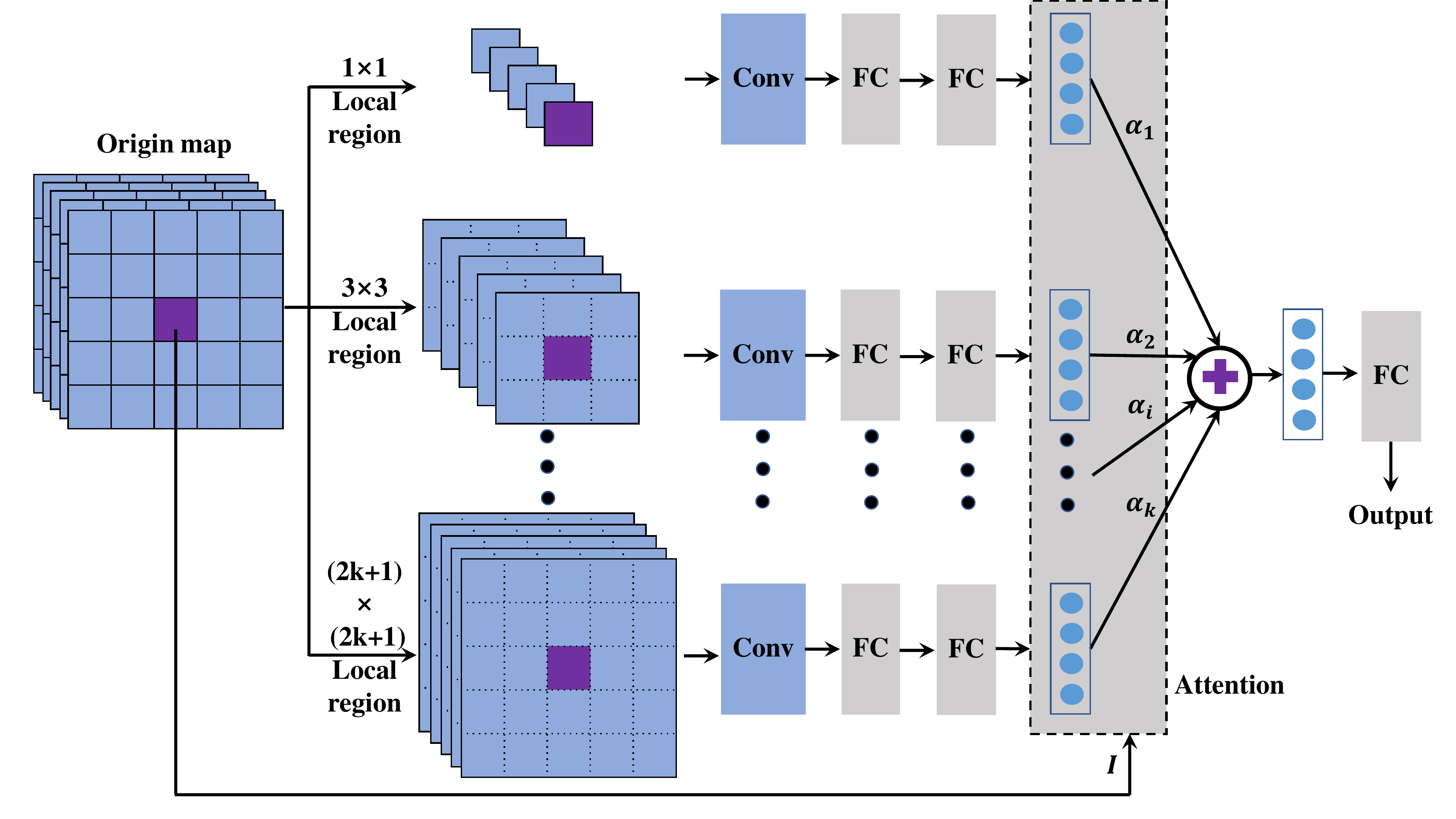}  
    \caption{The structure of our attention-based model. The depth of maps denotes the number of features of one grid. The purple grid is the target place of this sample for inferring bike demands. CNNs are applied to local regions of different sizes, i.e., $1 \times 1$, $3 \times 3$ and etc. The learned latent features from different local regions are merged with attention mechanism.}
    \label{fig:attModelStruc}
\end{figure*}

    With different local region sizes, we can produce several hidden vectors for each grid. The simplest way to merge these vectors is to concatenate them or compute average values for each dimension. However, the size of affecting areas of different grids may be different. For example, the size affecting areas of a metro station is larger than that of a park. In another word, different grids may prefer different sizes of local region in the bike demand inference task. Therefore, it may be not suitable to treat outputs of CNNs with different local region sizes equally or give them static weights.

    In this paper, we propose to utilize the attention mechanism to weight the hidden features from different local CNNs for each grid. The attention module is shown in Figure~\ref{fig:attModelStruc}. With the hidden features outputted by different local CNNs, we first compute the attention weight with the following equations, and we use the geo-related features $I_{g}$ of this grid $g$ and as one of the input.
    \begin{equation}\label{eq:att}
      \alpha_{i} = \frac{exp(I_{g} W^{'} Z_{i2})}{\sum_{j} exp (I_{g} W^{'} Z_{j2})},
    \end{equation}
    where $I_{g}$ is a feature vector with size equal to $d^{'}$  and $W{'}$ is a weight matrix.
    
    Then, the hidden features are merged with attention weights using element-wise product as shown in Equation~(\ref{eq:mergeFeature}).
    \begin{equation}\label{eq:mergeFeature}
      h = \sum_{i} \alpha_{i} Z_{i2},
    \end{equation}
    where $h$ is the result vector after attention mechanism.
    
    Finally, the attention module output a hidden vector for each grid as the representation. 
    With a fully connected layer, we infer fine-grained bike demand $\hat{d}_g$ for the grid, which is also the pattern on this grid.

\begin{algorithm}[b]
\SetKwInOut{Input}{\textbf{Input}}\SetKwInOut{Output}{\textbf{Output}} 
    \Input{
    \\
        POI features: $F_{p}, F_{p}^{'}$\;
        Road network features:  $F_{r},F_{r}^{'}$\;
        Satellite light features: $F_{s},F_{s}^{'}$\;
        Transportation features: $F_{t},F_{t}^{'}$\;
        Business centers: $F_{b},F_{b}^{'}$\;
        Bike demands: $D$, $D^{'}$ \; 
      }
    \Output{
        ALCNN model\;}
\verb|\\| Extract features and construct training instances\;
$F, F^{'} = coPCA ([F_{p}, F_{p}^{'}], [F_{r},F_{r}^{'}], [F_{s},F_{s}^{'}], [F_{t},F_{t}^{'}], [F_{b},F_{b}^{'}])$\; 

\For{all $g$ in city $S$}{
    $P(g) = DWT(D(g)) = DWT \left( [d_{g}^{X_1},d_{g}^{X_2},\cdots,d_{g}^{X_n}] \right)$\;
    Expand to local region: $F(g) \rightarrow T_{g}^{1\times1},T_{g}^{3\times3},T_{g}^{5\times5}, \dots$\;
    Put an  instance $(\left\{T_{g}^{1\times1},T_{g}^{3\times3},T_{g}^{5\times5}, \dots \right\}, P(g))$ in $\mathcal{D}$\;
}\nonl

\verb|\\| Train the model\;
Initialize all parameters $\theta$ in ALCNN\;
\Repeat {model converges or stopping criteria is met}{
    Randomly select a batch of instances $\mathcal{D}_{b}$ from $\mathcal{D}$\;
    Find $\theta$ by minimizing loss function KLMSE\;
}
\caption{ALCNN training process}
\label{al1}
\end{algorithm}

\subsection{Learning Process and Algorithm}

    To learn model parameters of our models, we use KLMSE as the objective function which is defined below:

    \begin{equation}
        KLMSE = \frac{1}{|G|} \sum_{g \in G} (KL(d_g , \hat{d}_g))^2,
    \end{equation}
    where $d_g$ is the corresponding ground truth, $\hat{d}_g$ is the inference of our model and $KL(\cdot)$ means the calculate of KL divergence. We adopt the mini-batch Adam to update parameters iteratively. To prevent overfitting in the training data, we apply dropout to randomly drop neurons in the two fully-connected layers in Equation~(\ref{layer}).   
    We also perform batch normalization to address the covariance shift problem and achieve faster convergence.
    
    Algorithm~\ref{al1} outlines the training process of our ALCNN. Notice that all features in the input contain two sets, one for source city and the other for target city. We first use coPCA and DWT to extract feature from multi-source geo-related data. For each grid, the features of its local region and its own bike demand pattern are used as one training sample. Next, we initialize our model. During the iterations, we randomly select a batch of training samples and update parameters using gradient descent until the model converges.


\section{EXPERIMENTS}\label{sec:experiment}

\subsection{Experimental Settings}

    \textbf{Datasets.} We managed to crawl Mobike data of three different cities, i.e., Beijing, Shanghai and Ningbo, China during 07/06/2017 and 15/07/2017. We also collected many kinds of geographic data, which are mentioned in Section~\ref{sec:featureExtract}. Table~\ref{tab:dataSta} shows the statistics of both our Mobike data and geographic data. 


\begin{table}[t]
    \centering
    \caption{Details of data}
    \label{tab:dataSta}
    \begin{tabular}{crrr}
    \toprule
    \multirow{1}*{Type} & \multicolumn{3}{c}{City} \\
    \cline{2-4}
            & Beijing & Shanghai & Ningbo\\
    \hline
    \multicolumn{4}{c}{\textbf{POI}}\\
    \# POIs             & 532,094   & 694,898                   & 85,613 \\
    POI categories      & 17        & 17    &  17 \\
    \hline
    \multicolumn{4}{c}{\textbf{Satellite light}}\\
    \# samples          & 23,021    & 28,954                    & 7,482 \\
    Average intensity(cd)   & 16.086 & 11.179  & 20.309\\
    Max distance(m)        & 18.779 & 32.305& 15.430 \\
    \hline
    \multicolumn{4}{c}{\textbf{Road networks}}\\
    \# roads            & 23,021 & 29,398 & 5,351 \\
    \# road levels      & 29 & 32  & 26\\
    \hline
    \multicolumn{4}{c}{\textbf{Transportation centers}}\\
    \# centers          & 334 & 366  & 53 \\
    Max distance(m)        & 33.475 & 43.526  & 10.861\\
    \hline
    \multicolumn{4}{c}{\textbf{Business centers}}\\
    \# centers          & 26 & 28 & 17 \\
    Max level           & 4 & 4 & 4\\
    \hline
    \multicolumn{4}{c}{\textbf{Mobike records}}\\
    \# bikes            & 656,437 & 591,295  & 35,591 \\
    Record amount       & 3,010,873 & 2,601,398  & 161,234\\
    \bottomrule
    \end{tabular}
\end{table}

    \textbf{Compared Methods.} As we study a novel research problem, there are few methods that are specially designed for solving it. As a result, we compare our proposed approach with several classic machine learning methods and one state-of-the-art approach.
    
    \begin{itemize}
        \item \textbf{Linear Regression (LR)}: This method ignores features of neighbors and infers bike demand of each grid independently using linear regression with $l2$-norm regularization.
        \item \textbf{KNN}. KNN predicts fine-grained bike demands by computing the average demands of $K$ nearest neighbors. The neighbors are selected from the training set using cosine similarity on geographic features.
        \item \textbf{RandomForest}. RandomForest constructs a multitude of decision trees to boost prediction performance.
        \item \textbf{XGBoost}~\cite{chen2016xgboost}. XGBoost is an optimized distributed gradient boosting method with high efficiency.
        \item \textbf{CoFA GeoConv}~\cite{liu2018inferring}. CoFA GeoConv is the state-of-the-art method for inferring bike demands and it combines joint Factor Analysis and convolutional neural network techniques. 
    \end{itemize}

    \textbf{Parameter Setting.} Generally, we tune all the methods mentioned before and report their performance with optimal parameter settings. LR, KNN, RandomForest, and XGBoost are implemented with the scikit-learn\footnote{\url{http://scikit-learn.org}} which is a popular Python machine learning library. For LR, we use $l2$ normalization and set penalty weight $\lambda = 0.1$. For KNN, we set the size of selected nearest neighbor to $k = 10$. For RandomForest, we use random state and bootstrap, and we set the number of estimators as $n_estimators = 500$.  For XGBoost, hyperparameters are set to $booster: gbtree$, $gamma = 0.2$, $max_depth = 14$, $lambda = 2$, $subsample = 0.9$, $colsample bytree = 0.7$, $min child weight = 4$, $\eta = 0.05$. As for CoFA GeoConv, we use the same parameters in their paper, i.e., $\lambda = 5$,  $\eta = 0.01$.
    
    We implement our approach with TensorFlow\footnote{\url{https://www.tensorflow.org/}}.
    We set the learning rate to $0.001$ and batch size to $128$. We also set our kennel size as $5 \times 5$, but if local region size is less or equal than $5 \times 5$, we will decline our kennel size correspondingly. We utilize the early stop method based on the performance on the validation set (stop until no increase in $50$ rounds). To prevent the model from overfitting, we employ dropout on the attention network and fully-connected layers and the dropout ratio is set to $0.1$. Batch normalization~\cite{ioffe2015batch} is conducted on the fully-connected layers for achieving faster convergence and better performance.
    
    \textbf{Evaluation Protocol.} As our task in this paper is to infer bike demands for assisting the deployment of bikes in a new city, we use the dataset of one city to train our model and transfer it to another city. We employ the KLMSE between the inference demands and the ground truth to evaluate the performance of various competing methods. Considering that the bikes are usually deployed first in developed cities rather than undeveloped cities, we test two kinds of transfer learning, i.e., transferring between two similar developed cities and transferring between a developed city and one not so developed city. In our experiments, we treat Beijing (BJ) and Shanghai (SH) as developed cities and Ningbo (NB) as a less developed city. We have three transfer learning tasks, including BJ $\rightarrow$ SH, SH $\rightarrow$ BJ, and SH $\rightarrow$ NB.

\subsection{Comparison Results}

\begin{table}[t]
    \centering
    \caption{Comparison among different methods}
    \label{tab:expCompare}
    \begin{tabular}{cccc}
        \toprule
        \multirow{1}*{Method} & \multicolumn{3}{c}{KLMSE} \\
        \cline{2-4}
                & BJ $\rightarrow$ SH & BJ $\rightarrow$ NB & SH $\rightarrow$ NB \\
        \hline
        LR & 0.232 & 0.506 & 0.493\\
        KNN  & 0.182& 0.213 & 0.201\\
        RandomForest  & 0.175& 0.199 & 0.187 \\
        XGBoost & 0.151 & 0.163 & 0.570 \\
        CoFA GeoConv  & 0.120& 0.135 & 0.129\\
        \hline
        ALCNN-DWT  & 0.107 & 0.120 & 0.117\\
        ALCNN-coPCA  & 0.104 & 0.116 & 0.113\\
        ALCNN-attention & 0.112 & 0.128 & 0.126\\
        ALCNN  & \textbf{0.093} & \textbf{0.104} & \textbf{0.099} \\
        \bottomrule
    \end{tabular}
\end{table}


    We first give the comparison with $5$ other models on the three transfer learning tasks. The comparison results are shown in Table~\ref{tab:expCompare}. ALCNN-DWT, ALCNN-coPCA, and ALCNN-attention denote our approach without DWT, coPCA, and attention modules, respectively.

    We find that our attention-based local CNN model performs best on all the three demand inference tasks and achieves the lowest KLMSE value. Among all baselines, LR performs the worst because it does not consider the influence of neighbors. Except LR, the KLMSE of other methods are all below $0.2$ as they all consider neighbors' features which proves the effectiveness of leveraging information of neighbors. RandomForest and XGBoost utilize multiple trees to perform ensemble learning which improves the inference performance. CoFA GeoConv achieves the lowest KLMSE among all baselines because it uses the coPCA to realize the distribution adaption between datasets of source city and target city. Except the coPCA, our ALCNN utilizes DWT and attention mechanism, and that's why our ALCNN performs better. Compared with ALCNN-DWT, ALCNN-coPCA, and ALCNN-attention, ALCNN with all the three modules achieves the best performance which proves the effectiveness of using DWT, coPCA and attention mechanism.

\subsection{Effectiveness of Attention Mechanism}

\begin{figure}[t]
    \centering
    \includegraphics[width = 0.4\textwidth]{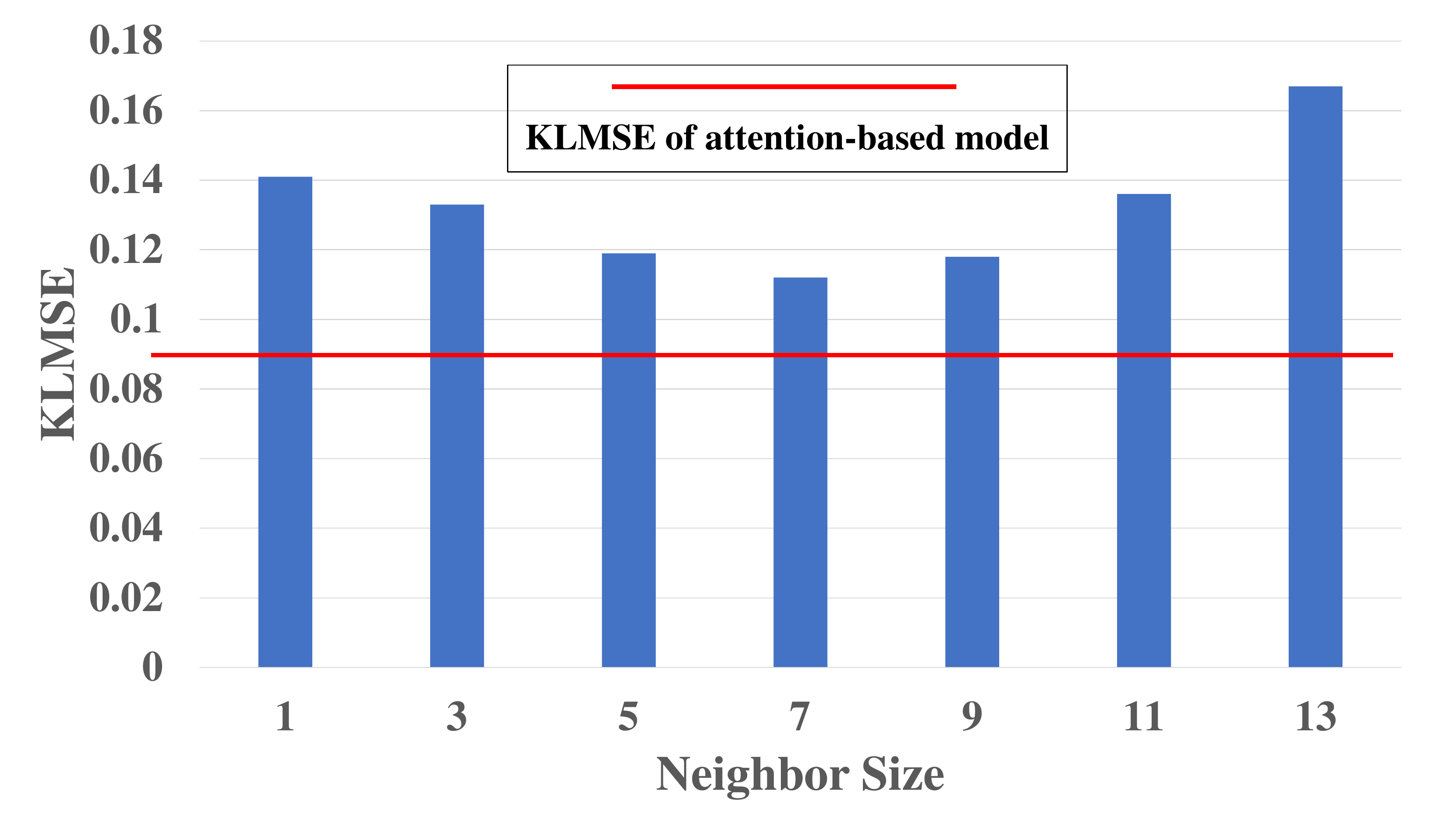}  
    \caption{Effectiveness of attention mechanism \\ (BJ $\rightarrow$ SH)}
    \label{fig:expAttention}
\end{figure}

    In our approach, the attention mechanism is aimed to automatically find the suitable local region size, i.e., determining the influence distance of neighbors. To further investigate whether the attention mechanism works, we conduct experiments to test the performance of our models with different \textbf{fixed} region sizes and compare them with the results of our method with attention mechanism. The experimental results are shown in Figure~\ref{fig:expAttention}.

    In the figure, we can see that our approaches with different sizes of local regions have different performances. When the local region size is $7 \times 7$, our approach achieves the lowest KLMSE. It proves that the local region size can affect the performance of the inference model and there exists one optimal local region size. In addition, we show the performance of our approach with attention mechanism with a red line in the figure. It shows that with the attention mechanism, our approach performs much better even compared with the method with the optimal fixed region size (i.e., $7$). We think the attention mechanism can select different optimal local region size for each grid which improves the inference performance, because different grids can affect areas of different sizes. E.g., the size of affecting areas of a metro station will be much larger than a park in the suburbans.

\subsection{Influence of Feature Extraction Module}

\begin{figure}[t]
    \centering
    \includegraphics[width = 0.4\textwidth]{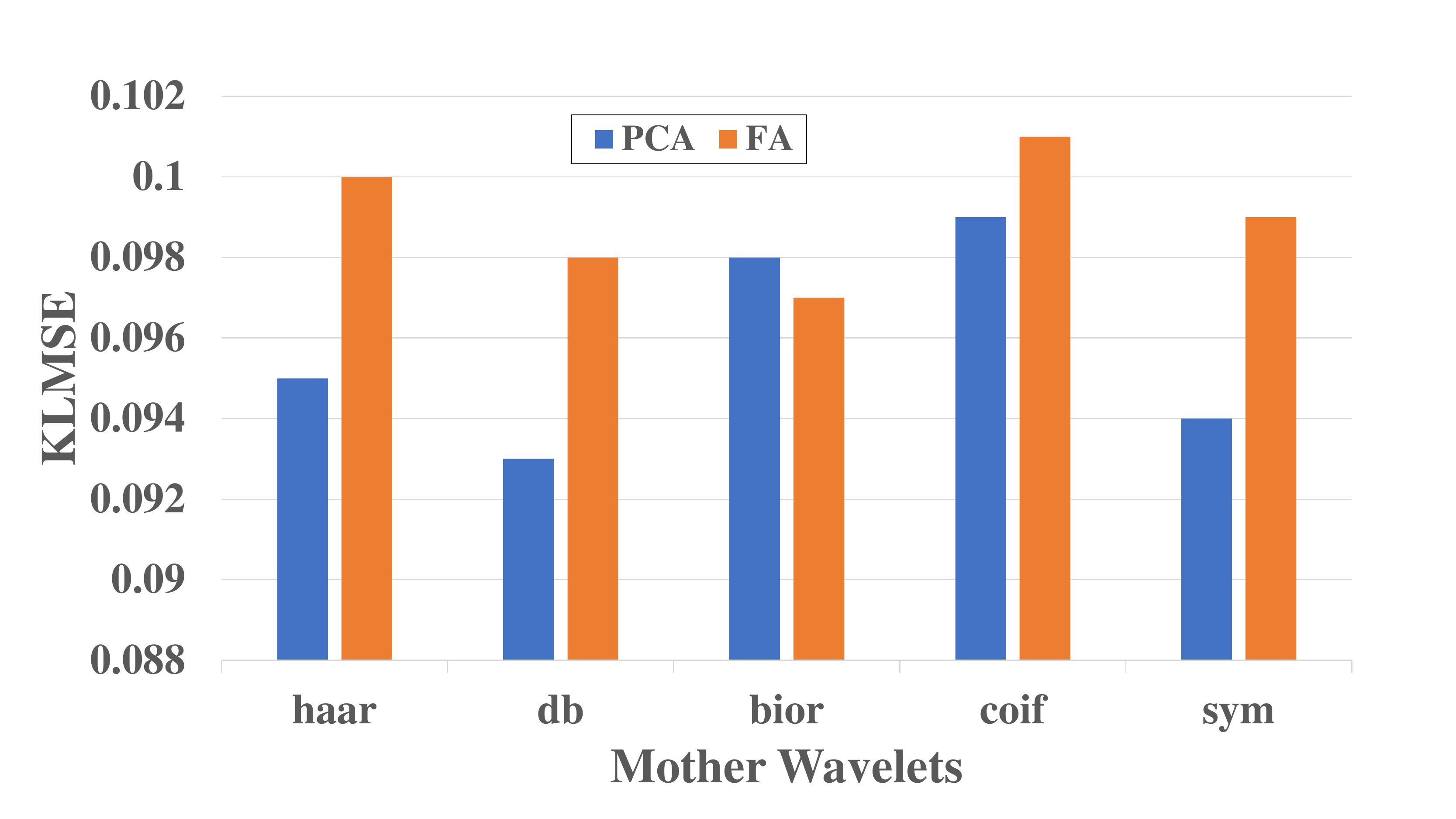}  
    \caption{Influence of our feature extraction model \\ \centerline{(BJ $\rightarrow$ SH)}}
    \label{fig:expFeature}
\end{figure}

    Our feature extraction module contains two parts: coPCA and DWT. The two parts realize the distribution adaption between datasets of source city and target city for input features and outputs, respectively, and improve the inference performance. Besides coPCA, there are some other methods can do the distribution adaption, including Factor Analysis (FA)~\cite{harman1960modern}. For DWT, there are many other wavelets to choose, like Haar wavelets and Daubechies wavelets. Therefore, we want to know if coPCA is better than FA and which wavelets are the best for DWT. We compare our approaches with different methods and the results are shown in Figure~\ref{fig:expFeature}.

    In the experiments, we will compare coPCA and FA, and test DWT with Haar wavelets (haar), Daubechies wavelets (db), Biorthogonal wavelets (bior),  Coiflets wavelets (coif), and Symlets wavelets (sym). We can see that all combinations of distribution adaption methods work good (KLMSE$<0.102$). When coFA is used, Biorthogonal wavelets turn out to work best, while when coPCA is used, Daubechies wavelets are leading. As we can see, Daubechies wavelets with coPCA has the lowest KLMSE, as a result, we decide to use Daubechies wavelets based DWT and coPCA for our feature extraction component.
    

\section{RELATED WORK}\label{sec:relatedWork}

\subsection{Research on Bike Sharing Systems and Urban Computing} 

    There are some researchers studying the problems in bike sharing systems. Some of them focused on the prediction of traffic flow based on historical data~\cite{ hoang2016forecasting,li2015traffic,yang2016mobility}. Zeng et al. introduced a station-level demand prediction method~\cite{zeng2016improving}. Other works mainly focused on the problem of station site optimization, which aimed to choose the best locations as bike stations sites~\cite{liu2015station,martinez2012optimisation}. However, they only focused on a single city populated with dock-less shared bikes, without applying the insights learned from a city into other cities. 
    
    Many works consider the temporal diversity in other fields of urban computing. Ferreira et al.'s work~\cite{ferreira2013visual} provides origin-destination queries from users that enable the study of mobility across the city, while Burns's work~\cite{burns2018temporal} provides an evaluating urban emissions in river system. The work influences us most is completed by Yao et al.~\cite{yao2018deep}, which proposes a multiple views of taxi demand pred view. However, none of these work solve the problem that the influence of an area is variable, which is the main contribution in our work.

\subsection{Transfer Learning on Urban Computing} 

    Some researchers applied transfer learning approaches for urban computing, especially the transfer learning among cities~\cite{pan2009survey}. Wei et al. solved
    the problem of predicting air quality in a target city by transferring knowledge learned from a source city~\cite{wei2016transfer}. Several approaches on domain adaptation~\cite{do2006transfer,dong2015multi,ganin2014unsupervised,ganin2016domain} have been proposed to adopt a model trained in the source domain to the target domain, some of which use the ideas of multi-view and multi-task. Among all related work, The work of Liu et al.~\cite{liu2018inferring} is the most related one to our work. They mainly focus on how to infer bike distribution in new cities, which is a predigestion of our problem. They propose a new way of feature exaction and fully apply the idea of transfer learning. Our work is greatly influenced by theirs in feature exaction, especially how to use geographic data. Since our problem is more complicated than theirs, we mainly focus on temporal patterns and attention mechanism. Their work doesn't consider the temporality of dock-less shared bikes, which is very important because dock-less shared bikes flow very frequently.  

\subsection{Attention Mechanism}

    Since the idea of attention mechanism was proposed, attention-based neural networks have been successfully used in many tasks~\cite{ba2014multiple,vaswani2017attention}, including many popular tasks: machine translation~\cite{luong2015effective}, computer version~\cite{you2016image}, speech recognition~\cite{chorowski2015attention} and healthcare~\cite{ma2018kame}. However, in the field of urban computing, the idea of attention mechanism is rarely used. Many works use the similar network as local CNN, but none of them add attention mechanism to it. Yao et al. point out local CNN may cause problems because the neighbor size is fixed~\cite{yao2018deep}, which does inspire us to try to use attention mechanism to find a suitable size of local neighbor for every grid.

\section{CONCLUSION}\label{sec:conclusion}

    In this paper, we mainly focus on the problem of inferring fine-grained daily bike demands in a new city, which is an important issue faced by bike sharing companies. We point out the importance to mine temporal patterns of fine-grained bike demands by DWT. We also adopt coPCA to extract features from multi-source geographic data which can be transferred to new cities. Then, we propose a local CNN to infer fine-grained bike demands with consideration of neighbors' influence. We also utilize attention mechanism to determine a suitable size of local region for every grid. The experiments on a real-world dataset demonstrate our proposed approach outperforms all competitive prediction methods and proves the effectiveness of the attention mechanism in the selection of local region sizes. 

%
%
%
\bibliographystyle{ACM-Reference-Format}
\bibliography{ref}
\citestyle{acmauthoryear}

%

\end{document}